%% file: main.tex
\documentclass[fleqn,10pt]{wlscirep}
\usepackage[utf8]{inputenc}
\usepackage[T1]{fontenc}
\usepackage{microtype}
\usepackage{lineno}
\usepackage[export]{adjustbox} 
\usepackage{standalone}
\usepackage{amsmath,amsfonts}
\usepackage{algorithmic}
\usepackage{algorithm}
\usepackage{array}
\usepackage{textcomp}
\usepackage{stfloats}
\usepackage{url}
\usepackage{verbatim}
\usepackage{listings}
\usepackage{graphicx}

\usepackage{pgfplots,pgfplotstable}
\usepgfplotslibrary{groupplots}
\pgfplotsset{compat=newest}

\usepackage[utf8]{inputenc}
\usepackage{tikz}
\usepackage[edges]{forest}  

\definecolor{folderbg}{RGB}{124,166,198}
\definecolor{folderborder}{RGB}{110,144,169}

\newlength\Size
\tikzset{%
  folder/.pic={%
    \filldraw [draw=folderborder, top color=folderbg!50, bottom color=folderbg] (-1.05*\Size,0.2*\Size+5pt) rectangle ++(.75*\Size,-0.2*\Size-5pt);
    \filldraw [draw=folderborder, top color=folderbg!50, bottom color=folderbg] (-1.15*\Size,-\Size) rectangle (1.15*\Size,\Size);
  },
  file/.pic={%
    \filldraw [draw=folderborder, top color=folderbg!5, bottom color=folderbg!10] (-\Size,.4*\Size+5pt) coordinate (a) |- (\Size,-1.2*\Size) coordinate (b) -- ++(0,1.6*\Size) coordinate (c) -- ++(-5pt,5pt) coordinate (d) -- cycle (d) |- (c);
  },
}

\forestset{%
  declare autowrapped toks={pic me}{},
  declare boolean register={pic root},
  pic root=0,
  pic dir tree/.style={%
    for tree={%
      folder, 
      font=\ttfamily,
      grow'=0, 
    },
    before typesetting nodes={%
      for tree={%
        edge label+/.option={pic me}, 
      },
      if pic root={
        tikz+={
          \pic at ([xshift=\Size].west) {folder};
        },
        align={l}
      }{},
    },
  },
  pic me set/.code n args=2{%
    \forestset{%
      #1/.style={%
        inner xsep=2\Size,
        pic me={pic {#2}}, 
      }
    }
  },
  pic me set={directory}{folder}, 
  pic me set={file}{file}, 
}

\usepackage{tumabbrev}
\usepackage{fourier}

\usepackage{enumitem}

\usepackage{xcolor,colortbl}
\definecolor{Gray}{gray}{0.85}
\definecolor{LightCyan}{rgb}{0.88,1,1}

\usepackage{siunitx}
\usepackage[capitalize,noabbrev,nameinlink]{cleveref} 
\usepackage{caption}
\usepackage{subcaption}
\renewcommand\thesubsection{\Alph{subsection}}
\renewcommand\thesubsubsection{\thesubsection.\arabic{subsubsection}}

\crefformat{footnote}{#2\textsuperscript{#1}#3}
\crefformat{enumi}{#2\textup{#1}#3}

\usepackage{tablefootnote}

\usepackage{multirow}
\usepackage{booktabs}
\usepackage{array}
\newcolumntype{x}{l}
\newcolumntype{v}[1]{>{\raggedright\hspace{0pt}}p{#1}}

\newcommand{\class}[1]{\texttt{\small#1}\xspace}
\newcommand{\meadow}{\class{pasture meadow grassland grass}}
\newcommand{\winterwheat}{\class{winter common soft wheat}}
\newcommand{\Winterwheat}{\class{Winter common soft wheat}}
\newcommand{\wheat}{\class{common soft wheat}}
\newcommand{\springwheat}{\class{spring common soft wheat}}

\newcommand{\dataset}[1]{\textsc{#1}\xspace}
\newcommand{\EuroCrops}{\dataset{EuroCrops}}
\newcommand{\EuroCropsML}{\dataset{EuroCropsML}}
\newcommand{\ZueriCrop}{\dataset{ZueriCrop}}
\newcommand{\Breizhcrops}{\dataset{BreizhCrops}}
\newcommand{\CropHarvest}{\dataset{CropHarvest}}
\newcommand{\Pastis}{\dataset{Pastis}}
\newcommand{\SenAgriNet}{\dataset{Sen4AgriNet}}
\newcommand{\DENETHOR}{\dataset{Denethor}}

\title{\EuroCropsML: A Time Series Benchmark Dataset For Few-Shot Crop Type Classification}

\author[1,*]{Joana Reuss}
\author[2,]{Jan Macdonald}
\author[2,3]{Simon Becker}
\author[2, 4]{Lorenz Richter}
\author[1]{Marco K\"orner}
\affil[1]{Technical University of Munich,
Chair of Remote Sensing Technology, Munich, 80333, Germany}
\affil[2]{dida Datenschmiede GmbH, Berlin, 10827, Germany}
\affil[3]{ETH Z\"urich, Department of Mathematics, Zürich, 8092, Switzerland}
\affil[4]{Zuse Institute, Berlin, 14195, Germany}

\affil[*]{corresponding author: Joana Reuss (joana.reuss@tum.de)}

\begin{abstract}
We introduce \EuroCropsML, an analysis-ready remote sensing machine learning dataset for time series crop type classification of agricultural parcels in Europe. 
{It is the first dataset designed to benchmark transnational few-shot crop type classification algorithms} that supports advancements in algorithmic development and research comparability. 
It comprises \num{706683} multi-class labeled data points across \num{176} classes, featuring annual time series of per-parcel median pixel values from Sentinel-2 L1C data for the year 2021, along with crop type labels and spatial coordinates. 
Based on the open-source \EuroCrops collection, \EuroCropsML is publicly available on Zenodo: \url{https://zenodo.org/records/13789558}.
\end{abstract}
\begin{document}

\flushbottom
\maketitle

\thispagestyle{empty}

\section*{Background \& Summary}
The availability of spatio-temporal satellite imagery and the success of data-driven modeling lead researchers to explore \emph{machine learning (ML)} methods for diverse remote sensing tasks.

One key application is monitoring agricultural crop distribution worldwide, which is crucial for food security, a core \emph{sustainable development goal (SDG)} set by the \emph{United Nations (UN)}. 
With five billion hectares of agricultural land globally, \cite{land-use} \emph{crop type classification} has become a central focus in remote sensing. Consequently, numerous datasets have been compiled, such as \ZueriCrop~\cite{turkoglu_cropmapping} for northern Switzerland, \DENETHOR~\cite{DENETHOR}, a non-publicly available dataset for northern Germany, \Breizhcrops~\cite{russwurm_breizhcrops} for the French Brittany region, \Pastis~\cite{Garnot2021PanopticSO} also covering France, and \SenAgriNet~\cite{Dimitris_2021_sen4agrinet} covering France and the Spanish autonomous community Catalonia. Further, there is the \CropHarvest~\cite{tseng_cropharvest_2021} collection, which is a global dataset mainly featuring binary crop-\vs-non-crop labels. For a comprehensive listing, we also refer to a recent overview~\cite{Schmitt_2023} over different available datasets focusing more broadly on earth observation data. 

However, as shown in \cref{tab:comparison}, current crop classification datasets are commonly restricted to small areas within a single country. They comprise only a small number of multi-class crop labels, or include a limited number of agricultural parcels. This hinders the effective benchmarking of data-driven methods.
With \EuroCropsML, we aim to address the shortcomings of existing datasets and provide the first comprehensive dataset that covers many desirable aspects for benchmarking ML algorithms, with an emphasis on transfer and, in particular, few-shot learning.

\emph{Transfer learning}, instead of directly training a model from scratch on limited target data, uses model parameters obtained from pre-training on related data as a starting point for fine-tuning on the target task.
\emph{Few-shot learning} algorithms, on the other hand, aim to enable models to perform well on tasks with very few training examples to quickly adapt to new scenarios not represented in the training data.
Thus, crop type classification can benefit by leveraging detailed ground-truth data from well-documented countries to enhance classification in regions with only limited data available, potentially across varying climates and crop types.

\emph{Supervised learning} algorithms for crop type classification from remote sensing data require high-quality class label annotations in addition to the satellite imagery. The open-source \EuroCrops dataset \cite{schneider_eurocrops_zenodo,schneider_eurocrops_2023} provides data on agricultural parcels and crop types across the \emph{European Union (EU)}. 
The data has been collected directly from farmers' self-declarations. 
Its harmonized \emph{hierarchical crop and agriculture taxonomy (HCAT)} \cite{schneider_eurocrops_2021,schneider_eurocrops_2023} makes it a valuable resource for developing and benchmarking ML algorithms across countries. 
The hierarchically structured and harmonized crop classes facilitate the identification of closely related crops.
For instance, all cereals and flowers are distinguished from each other at a higher level: The taxonomy assigns a greater proximity to wheat and oats (both cereals) than to wheat and tulips (flower).
This is advantageous for interpreting algorithmic performance and learning beyond binary crop-\vs-non-crop labeling.
Our dataset \EuroCropsML combines Sentinel-2 L1C time-resolved multi-spectral satellite imagery data from the year 2021 with the harmonized crop type class label annotations obtained from the \EuroCrops reference dataset. 

\begin{figure}
    \centering
    \begin{minipage}[t]{6.8cm}
        \vspace{0pt}
        \raggedright
        \includegraphics[height=7.5cm, keepaspectratio, valign=t]{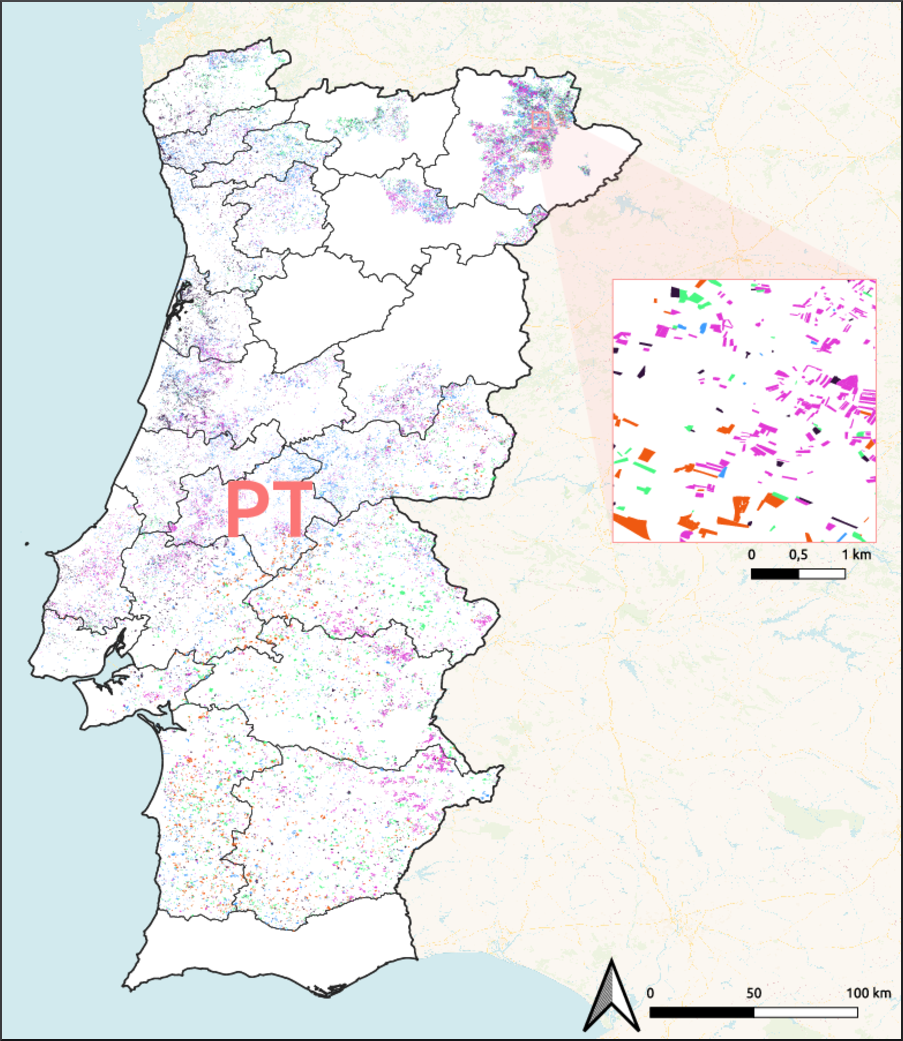}
    \end{minipage}
    \begin{minipage}[t]{8cm}
        \vspace{0pt}
        \raggedright
        \includegraphics[height=6cm, keepaspectratio, valign=t]{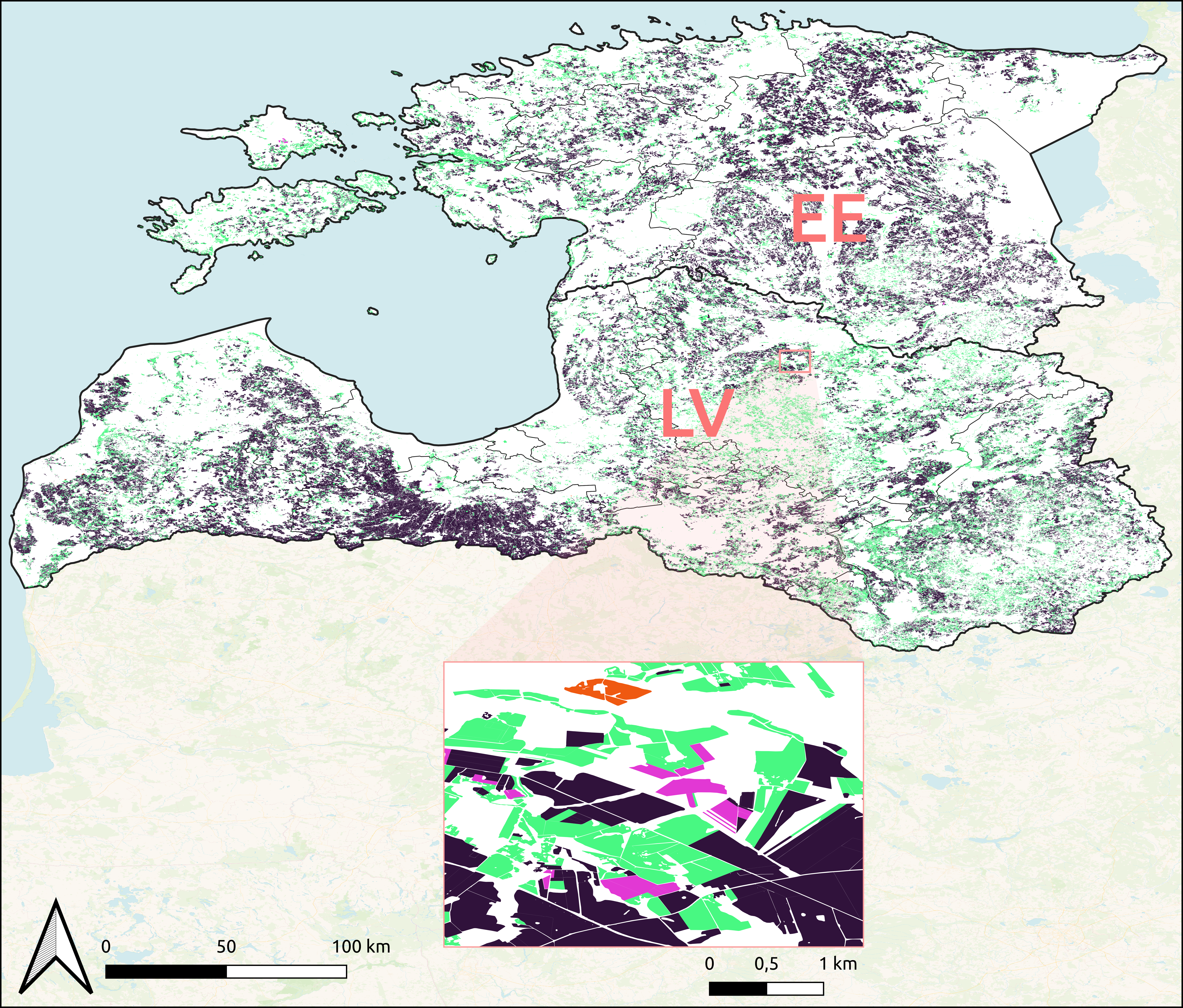}\\[2.5mm] 
        \includegraphics[height=1.1cm, keepaspectratio, valign=b]{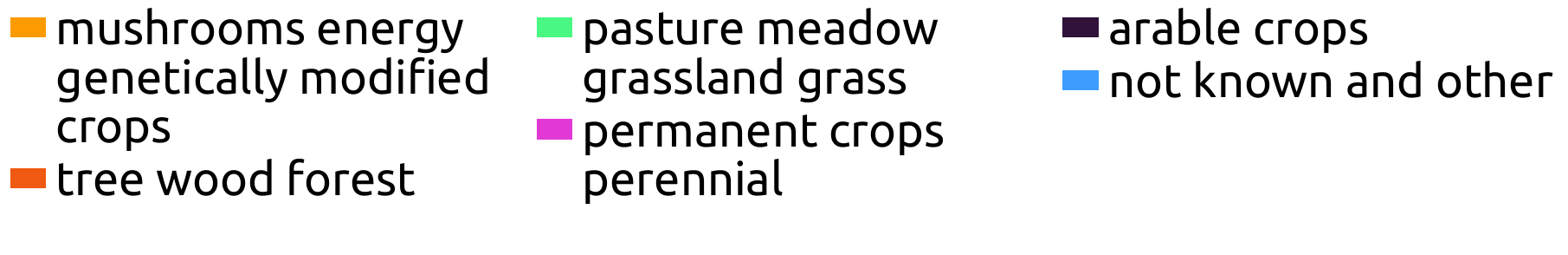} 
    \end{minipage}
    \caption{Visualization of crop fields (using \EuroCrops HCAT3 level \num{3} \cite{schneider_eurocrops_2021, eurocrops_github}) for Estonia (EE), Latvia (LV), and Portugal (PT). 
    The majority of Latvia and Estonia is comprised of expansive clusters of \class{arable crops} and \class{pasture meadow grassland grass}. 
    In contrast, Portugal presents a more diverse distribution among the various classes, with crop fields exhibiting a more dispersed pattern across the country.}
    \label{fig:eurocropsmlmap}
\end{figure}

\begin{table}
    \centering
    \begin{tabular}[t]{@{}xlrrrr@{}}
    \toprule
    Dataset & \multicolumn{1}{x}{Transnational} & \multicolumn{1}{x}{\# Classes} & \multicolumn{1}{x}{Area (in \unit{\kilo\meter\squared})} & \multicolumn{1}{x}{\# Parcels} & \multicolumn{1}{x@{}}{\# Time steps}\\
    \cmidrule(r){1-1} \cmidrule(lr){2-2} \cmidrule(lr){3-3} \cmidrule(lr){4-4} \cmidrule(lr){5-5} \cmidrule(l){6-6}
    \ZueriCrop \cite{turkoglu_cropmapping} & {no} & {\num{48}} & {\num{2400}} & {\num{116000}} & {\color{teal}\num{71}} \\
    \Breizhcrops \cite{russwurm_breizhcrops} &{no}& {\num{9}} & {\num{27200}} & {\color{teal}\num{768175}} & {\color{teal}\num{51}--\num{102}}\\
    \CropHarvest \cite{tseng_cropharvest_2021} &{\color{teal}yes} & {\text{N/A}$^{\dagger}$} & {\color{teal}\text{global}} & {\num{90480}} & {\num{12}}\\
    \Pastis \cite{Garnot2021PanopticSO} & {no} & {\num{18}} & {\num{4000}}  & {\num{124000}} & {\num{1}} \\
    \DENETHOR \cite{DENETHOR} & {no} & {\num{9}} & {Northern Germany} & {4500} & {\color{teal} daily} \\
    \SenAgriNet \cite{Dimitris_2021_sen4agrinet} & {\color{teal}\text{yes}}& {\color{teal}\num{168}} & {France \& Catalonia} & {\color{teal}\num{42.5} mio} & {\num{5}}\\
    \textbf{\EuroCropsML} &{\color{teal}yes} & {\color{teal}\num{176}} & {\color{teal}\num{222110}} & {\color{teal}\num{706683}} & {\color{teal}\num{1}--\num{216}}\\
    \bottomrule
    \end{tabular}
    \caption{Comparison between \EuroCropsML and other datasets with respect to availability of transnational data, number of crop type classes, covered area, number of parcels, and number of time step observations per data point. 
    Suitable features that \textcolor{teal}{qualify} the dataset for real-world benchmarking purposes are colored.\\
    $^{\dagger}$ The \CropHarvest class labels are not standardized between regions, \SI{65.8}{\percent} of \CropHarvest data is only binary labeled (crop vs. non-crop),  it contains \num{348} unstructured classes in total.
    }
    \label{tab:comparison}
\end{table}

In summary, the strengths of our dataset are:
\begin{enumerate}
    \item Fully annotated remote sensing crop type classification data with a coverage of diverse geographical regions representing different climate zones as well as different vegetation and agricultural practices. 
    \item Self-declared multi-class labels for \num{706683} data points classified into \num{176} distinct crop classes, refering to the final data after pre-processing, \cf \cref{sec:level2_data_processing} in the Methods section.
    This addresses a significant limitation of some existing datasets, where labels may be less accurate and streamlined. 
    \item Harmonized crop labels across multiple countries that enable in-depth analyses of knowledge transfer between geographical regions based on crop taxonomy. 
    \item Fine-grained annual time series of up to \num{216} time steps per data point.
    Again, this refers to the final data after pre-processing, \cf \cref{sec:level2_data_processing} in the Methods section.
    \item Provisioning of pre-built training and evaluation dataset splits ready-to-use for benchmarking crop type classification ML algorithms.
\end{enumerate}

\section*{Methods}
\subsection{Country selection}
The \EuroCropsML dataset comprises three \emph{regions of interest (ROIs)}.
They are selected based on the availability of sample data points and to showcase locations with both similar and different climates, vegetation, and crop classes. 
The ROIs are Estonia, Latvia, and Portugal, as shown in \cref{fig:eurocropsmlmap}. 
Latvia and Estonia are neighboring countries. 
In contrast, Portugal is located in a different climate zone, with different vegetation and cultivation practices, \cf \cref{fig:eurocropsmlmap}.
Furthermore, \cref{fig:areasamples} shows the disparities in parcel size, another notable distinction between the countries' cultivation practices.
All three countries collectively provide a substantial number of overlapping and non-overlapping crop classes (\cf \cref{fig:numbersamples}). 
The dataset reflects the high class imbalances common in real-world crop type classification.
For instance, the crop type class \meadow is by far the largest class in the considered ROIs, accounting for roughly \SI{45}{\percent} of all data points (\SI{48}{\percent} in Estonia, \SI{50}{\percent} in Latvia,  \SI{17}{\percent} in Portugal).

\begin{figure}
    \centering
    \begin{subfigure}[b]{0.49\columnwidth}
        \resizebox{.91\linewidth}{!}{\input{images/area_histograms}}
        \caption{\label{fig:areasamples} Number of parcels (with a log scale) with a certain area}
    \end{subfigure}
    \begin{subfigure}[b]{0.49\columnwidth}
        \resizebox{.95\linewidth}{!}{\input{images/class_histograms}}
        \caption{\label{fig:numbersamples} Number of parcels (with a log scale) with a certain crop class}
    \end{subfigure}
    \caption{\textbf{(a)} Number of parcels (with a log scale) of a certain size within Estonia, Latvia, Portugal, and the overall \EuroCropsML dataset. The histogram bin width for the parcels sizes is \SI{0.25}{\kilo\meter\squared}. \textbf{(b)} Number of parcels (with a log scale) of all \num{176} distinct \EuroCropsML crop classes (HCAT3 level \num{6} \cite{schneider_eurocrops_2021,schneider_eurocrops_2023}) within Estonia, Latvia, Portugal, and the overall \EuroCropsML dataset. The crop class with---by far---the largest prevalence is the meadow class.}
    \label{fig:parcels_distribution}
\end{figure}

\subsection{Data collection}
\begin{figure}
\centering
\includegraphics[width=0.95\linewidth]{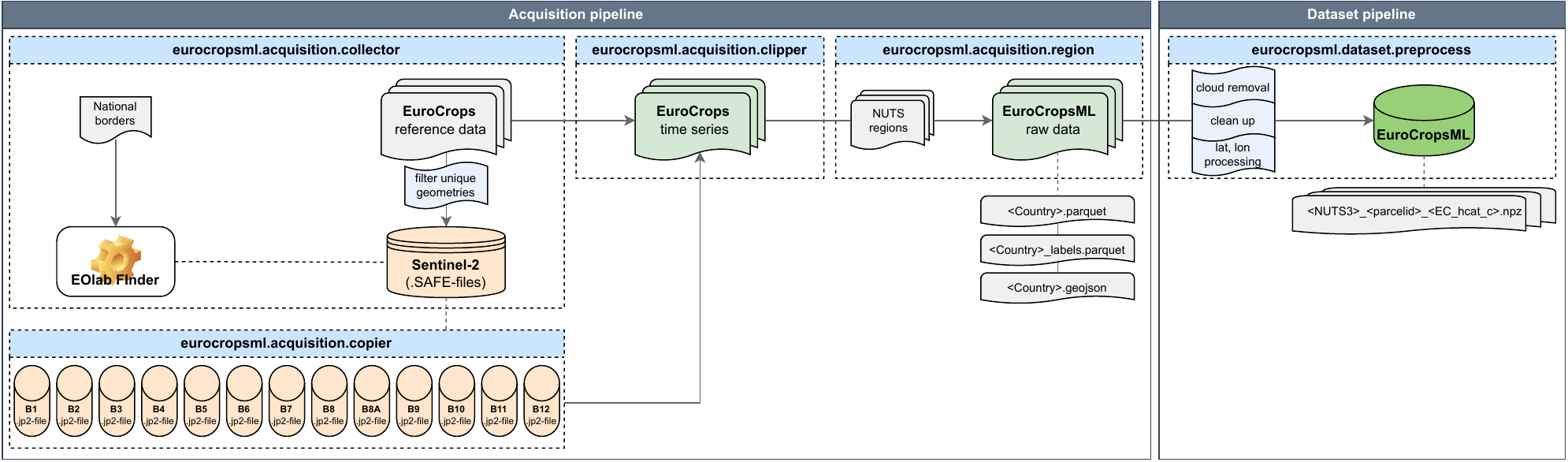}
\caption{Overview of the data acquisition and pre-processing pipeline for the \EuroCropsML dataset. 
The names in the blue headers correspond to the location and module names of the respective step in the associated Python package, available at \url{https://github.com/dida-do/eurocropsml} (\cf \cref{sec:level1_usage,sec:level2_usage} in the Usage Notes).}
\label{fig:pipeline}
\end{figure}

The country-specific data for \EuroCropsML is generated through the process illustrated in \cref{fig:pipeline} and can be broken down into two main sub-processes: data acquisition and pre-processing.
Data points in the dataset correspond to one agricultural parcel and are represented by an annual time series of multi-spectral Sentinel-2 observation data.
More precisely, a single observation contains the median pixel values across the parcel's area for each of the \num{13} Sentinel-2 spectral bands and each of the observed points in time. \cite{schneider_eurocrops_2021}

\subsubsection{Level I: Data acquisition}
\label{sec:level1_data_collection}
\begin{figure}
\centering
\includegraphics[width=0.95\linewidth]{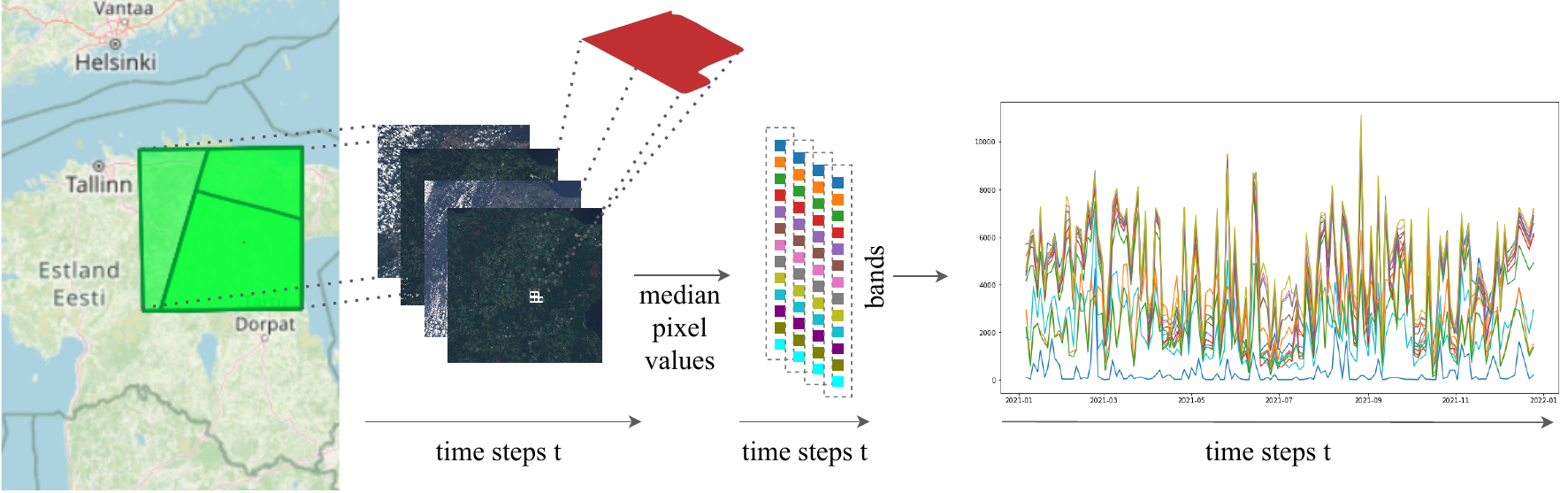}
\caption{Illustration of the data processing to obtain a median pixel time series from Sentinel-2 raster tiles for a single agricultural parcel. 
All Sentinel-2 tiles for the year 2021 that overlap with the parcel's geometry (red polygon) are collected. 
For each individual time step and Sentinel-2 band, they are clipped to the extent of the polygon before calculating the median pixel value for each time step and band individually, resulting in a multi-spectral time series.}
\label{fig:clipping}
\end{figure}

\begin{description}
    \item[Collection of relevant Sentinel-2 tiles.] In the first step, the \EuroCrops reference data \cite{schneider_eurocrops_zenodo} (version 9) was aligned with Sentinel-2 raster data tiles from the year 2021.
    For this, we relied on the EOLab Finder, accessible via \url{https://finder.eo-lab.org}, to identify the file paths of the Sentinel-2 \texttt{\small .SAFE} files on the EOLab platform (\url{https://eo-lab.org/de}).
    Please note that the EOLab Finder is no longer maintained since May 11, 2024.
    As a result, when processing future data (\cf \cref{sec:level1_usage} in the Usage Notes section), please refer to its successor, the EOLab Data Explorer, accessible via \url{https://explore.eo-lab.org}, instead.
    We first collected all tiles that overlap with the land surface of the given country for 2021, before mapping the collection to each agricultural parcel individually.
    It should be noted that some \EuroCrops reference data contains duplicate parcel geometries.
    In such instances, only one entry is retained.
    Furthermore, if a parcel encompasses multiple raster tiles, only the tile with the lowest cloud coverage was retained for further processing.
    Thus, in such rare cases, only parts of the parcel's geometry were included in the polygon clipping and median pixel value calculation described in following step. 
    However, since all subsequent processing and modeling steps relied solely on median pixel values rather than on individual pixels, merging information from multiple tiles is unnecessary.
    
    \item[Clipping of satellite data and calculation of median pixel values.]
    Parcels were extracted from the dataset to isolate the satellite images corresponding to each geographical unit.
    The Sentinel-2 data provides satellite imagery with high spatial resolutions of \qtylist{10;20;60}{\meter} per pixel, depending on the spectral band. 
    The combined Sentinel-2 constellation has a revisit time of five days, resulting in a temporal resolution that allows for a significant amount of observations for each agricultural parcel throughout the year.
    For every parcel and at each available time step, we calculated the median pixel value for each of the \num{13} spectral bands in the Sentinel-2 raster tiles.
    This provides a time series containing optical observations for each parcel. 
    It quantifies the light reflected by the Earth's surface across various wavelengths over the duration of a year.
    Our derivation of median pixel time series from Sentinel-2 raster tiles is illustrated in \cref{fig:clipping}.
    
    \item[Regional mapping.]
    For improved geographical accuracy and effective spatial sub-division of the dataset, we employed the Eurostat's \emph{GISCO} database to associate the \EuroCrops parcels with their corresponding \emph{nomenclature of territorial units for statistics (NUTS)} region.
    These regions are organized in a hierarchical system from level 1 (large socio-economic regions) to level 3 (detailed local areas). This mapping process generated the \emph{raw} \EuroCropsML \emph{dataset} (\cref{stage1} in the Data Records section), which incorporates parcel geometries, crop type label information, and the associated Sentinel-2 time series.
\end{description}

\subsubsection{Level II: Data pre-processing}
\label{sec:level2_data_processing}
Additional pre-processing steps were performed to convert the {raw \EuroCropsML} dataset (\cref{stage1} in the Data Records section) into a ready-to-use ML dataset (\cref{stage2} in the Data Records section).
\begin{description}
    \item[Cloud removal.] 
    We performed a cloud removal step following the scene classification approach of the Level-2A algorithm. \cite{ESA_Sentinel2_ATBD}
To identify clouds, we relied on brightness thresholds for the red spectral band (B4).
If the median reflectance of the band fell below a prescribed threshold ($t_1=0.07$), it was considered cloud-free and deterministically assigned a cloud probability of \qty{0}{\percent}.
Conversely, if the value exceeded another prescribed threshold ($t_2=0.25$), it was considered cloudy and we assigned a cloud probability of \qty{100}{\percent}.
All values between the aforementioned thresholds were linearly interpolated and assigned probabilities between \qty{0}{\percent} and \qty{100}{\percent}.
Consequently, all observations with a cloud probability greater than \qty{50}{\percent} ($p=0.5$) were removed. 
Data points that only contained time series points with a cloud coverage score greater than \qty{50}{\percent}, were removed from the dataset completely.
    \item[File name convention and metadata.]
To facilitate data loading during the training of ML models, each data point is stored separately as a \texttt{\small NumPy} \texttt{\small .npz} file, accompanied by metadata, such as the observations timestamps and the spatial coordinates of the parcel's centroid.
The naming convention \texttt{\small <NUTS3-region>\_<parcelID>\_<EC\_hcat\_c>.npz} is used for all \texttt{\small .npz} files, where \texttt{\small EC\_hcat\_c} is the \EuroCrops HCAT3 crop class code \cite{schneider_eurocrops_2021,schneider_eurocrops_2023} (\cf the Data Records section).
\end{description}

\subsection{Benchmark tasks}
\label{sec:benchmkaring_tasks}
A main goal of \EuroCropsML is to facilitate research focused on knowledge transfer between different geographical regions. 
We propose two transfer learning scenarios by splitting the dataset into subsets for training (\cref{stage3} in the Data Records section). 
These are crucial for evaluating and optimizing ML model performance in crop classification. The data splits are:
\begin{description}
    \item[Latvia$\rightarrow$Estonia (LV$\rightarrow$EE)] The models are pre-trained on data from Latvia only and then fine-tuned and evaluated on data from Estonia.
    \item[Latvia+Portugal$\rightarrow$Estonia (LV+PT$\rightarrow$EE)] The pre-training
    is conducted on data from Latvia and Portugal, followed by fine-tuning and evaluation on the same data from Estonia as in the first scenario.
\end{description}
{The two cases are intentionally selected to benchmark algorithms on their performance in geographically and agriculturally distinct and similar regions.}
For the pre-training stage, we downsample the class \meadow to the median frequency of all other classes.
For fine-tuning, we create eight different few-shot learning scenarios simulating challenging real-world use cases of varying data scarcity. 
In addition to utilizing all available data from the fine-tuning training subset, we provide splits for 1-, 5-, 10-, 20-, 100-, 200-, and 500-shot classification. 
That is, for a $k$-shot split, for each class in the training set of the fine-tuning data, a maximum of $k$ samples is used (or as many as there are available if this is fewer than $k$).
The process of generating the splits can be outlined as:
\begin{description}
    \item[Pre-train dataset] \hfill
        \hspace{0pt}
        \begin{enumerate}[noitemsep,nolistsep]
            \item Filtering pre-processed files for relevant pre-training NUTS-regions
            \item Downsampling \meadow class
            \item Creating the training-validation split using \texttt{\small scikit-learn}
        \end{enumerate}
    \item[Fine-tune dataset]\hfill
        \hspace{0pt}
        \begin{enumerate}[noitemsep,nolistsep]
            \item Filtering pre-processed files for relevant fine-tuning NUTS-regions 
            \item Creating the training-validation-testing split using \texttt{\small scikit-learn}
            \item Sampling \num{1000} samples from the validation subset
            \item Sampling of $k \in \{1,5,10,20,100,200,500\}$ samples per class from the training subset
        \end{enumerate}
\end{description}

\section*{Data Records}
\label{sec:datarecords}
The \EuroCropsML dataset is publicly hosted on Zenodo. \cite{reuss_macdonald_eurocropsml_2024}
We provide three processing stages of the dataset:
\begin{enumerate}[label=S.\arabic*, font=\bfseries]
    \item \label{stage1}\textbf{raw data}: The raw Sentinel-2 L1C time series, including all available observations, allowing researchers to perform their own pre-processing steps, \cf \cref{sec:level1_data_collection} in the Methods section.
    \item \label{stage2}\textbf{pre-processed data}: The ready-to-use ML dataset after cloud removal, additional data cleaning, and calculation of spatial coordinates of the parcel's centroid, \cf \cref{sec:level2_data_processing} in the Methods section.
    \item \label{stage3}\textbf{pre-built splits}: Multiple pre-training (training and validation) and fine-tuning (training, validation, and testing) task configurations for benchmarking few-shot crop type classification (\cf \cref{sec:benchmkaring_tasks} in the Methods section) in a transfer learning scenario. 
    This is a decomposed version of \cref{stage2} and must be used in combination with the pre-processed data.
\end{enumerate}

\cref{fig:zenodotree} shows the structure of the data records on Zenodo, wherein each of the aforementioned stages is represented by a distinct directory.
The dataset is organized into individual sub-collections for each country, facilitating the future addition of additional countries.

\begin{figure}
    \centering
    \begin{subfigure}[t]{0.31\linewidth}
        \centering
        \vspace{0cm}
        \resizebox{!}{6.5cm}{\input{images/zenodo/zenodorawdata}}
        \vspace{2.15cm}
        \caption{\label{fig:zenodorawdata} Raw data (\cf \cref{sec:level1_data_collection} in the Methods section and \cref{stage1} in the Data Records section). The files designated as \texttt{\small <Country>.parquet} contain the actual cropped time series and reference data. Furthermore, the files named \texttt{\small <Country>\_labels.parquet} contain only the parcel-label matches, thus enabling rapid loading of the labels.}
    \end{subfigure}
    \hspace{0.01\linewidth}
    \begin{subfigure}[t]{0.34\linewidth}
        \centering
        \vspace{0cm}
        \resizebox{!}{6.9cm}{\input{images/zenodo/zenodopreprocess}}
        \vspace{1.75cm}
        \caption{\label{fig:zenodopreprocess} Pre-processed and ready-to-use dataset for ML (\cf \cref{sec:level2_data_processing} in the Methods section, and \cref{stage2} in the Data Records section).}
    \end{subfigure}
    \hspace{0.01\linewidth}
    \begin{subfigure}[t]{0.31\linewidth}
        \centering
        \vspace{0cm}
        \resizebox{!}{8.5cm}{\input{images/zenodo/zenodosplit}}
        \vspace{0.15cm}
        \caption{\label{fig:zenodosplit} Pre-defined data splits used for benchmarking (\cf \cref{sec:benchmkaring_tasks} in the Methods section, and \cref{stage3},  in the Data Records section).}
    \end{subfigure}
    \caption{Data records in Zenodo, referring to version 8 of the dataset.}
    \label{fig:zenodotree}
\end{figure}

\section*{Technical Validation}
As the ground truth label annotations and parcel geometries are directly sourced from \EuroCrops, \cite{schneider_eurocrops_2023} our validation in the following is limited to the processing of the Sentinel-2 data as well as the benchmarking feasibility of the proposed dataset.
As there are no Sentinel-2 ground truths available for the time series, alternative measurements were taken to validate the data processing and serve as sanity checks on several data samples.

\subsection{Clipping process}
Once the clipped median values for each country had been obtained, it was ensured that each parcel had values present.
Thereafter, we verified the average \emph{normalized difference vegetation index (NDVI)} per class and country.
The NDVI is a measure of the amount of light reflected by a plant.
It is therefore employed in the monitoring of crop growth periods, with higher NDVI values indicating greater crop biomass, and can serve as an instrument for verifying the accuracy of Sentinel-2 time series.
Consequently, during the flowering season, we anticipate to observe higher NDVIs, whereas during the planting season, we expect values close to zero.
\Cref{fig:ndvi}, for instance, shows the mean annual NDVI for \winterwheat/\springwheat and \wheat per country.
\Winterwheat in Latvia and Estonia is planted before the winter season.
The NDVI shows an initial increase during the early stages of spring, reaching its highest values around the onset of summer, which coincides with its flowering season.
In contrast, \springwheat is planted in the early stages of spring.
Consequently, the NDVI demonstrates a slight delay in its increase in comparison to that observed in \winterwheat. 
In general, \winterwheat exhibits slightly higher NDVI peaks than its sibling \springwheat.
This is attributable to its head start and the longer growing period it enjoys.
However, both crops exhibit analogous biological phases upon entering active growth, resulting in their peaks occurring concurrently.

\begin{figure}
    \centering
    \begin{subfigure}[t]{0.345\linewidth}
        \centering
        \vspace{0cm}
        \resizebox{!}{4.5cm}{\input{images/NDVI/NDVI_LV}}
        \caption{\label{fig:lv_ndvi} Latvia}
    \end{subfigure}
    \hspace{0.001\linewidth}
    \begin{subfigure}[t]{0.3175\linewidth}
        \centering
        \vspace{0cm}
        \resizebox{!}{4.5cm}{\input{images/NDVI/NDVI_EE}}
        \caption{\label{fig:ee_ndvi} Estonia}
    \end{subfigure}
    \hspace{0.001\linewidth}
    \begin{subfigure}[t]{0.3175\linewidth}
        \centering
        \vspace{0cm}
        \resizebox{!}{4.5cm}{\input{images/NDVI/NDVI_PT}}
        \caption{\label{fig:pt_ndvi} Portugal}
    \end{subfigure}
    \caption{Mean annual NDVI for the crop type \winterwheat/\springwheat for Latvia and Estonia and \wheat for Portugal after cloud removal.
    It should be noted that Portugal does not distinguish between winter and spring wheat.
    Hence, the mean NDVI may encompass both classes.
    }
    \label{fig:ndvi}
\end{figure}

\subsection{Cloud removal}
The cloud removal step is derived from the official scene classification approach of the Level-2A algorithm. \cite{ESA_Sentinel2_ATBD} Consequently, it has already been validated.
However, as only a subset of the complete approach was employed, the method was validated on our data by visualizing the Sentinel-2 time series before and after cloud removal (\cf \cref{fig:reflectance}).

\begin{figure}
    \centering
    \resizebox{.99\linewidth}{!}{\input{images/reflectance/reflectance}}
    \caption{Sentinel-2 reflectance data for an exemplary sample of  \winterwheat in Estonia (showing the three visible light bands (B02 blue, B03 green, B04 red) in their respective color and the remaining bands (B01 aerosol, B05--B8A VNIR, B09--B12 SWIR) in different shades of pink and purple).
    The high peaks present in the left plot are caused by cloudy observations and are removed in the right plot after applying the cloud removal step.}
    \label{fig:reflectance}
\end{figure}

\subsection{Benchmarking}
The integrity of the benchmarking process on our dataset is guaranteed by ensuring that the data intended for the pre-training and fine-tuning stages do not include any overlapping samples.
Furthermore, we also ensured that in all cases the pre-defined data split subsets (training/validation/testing) are disjoint.

We demonstrate the benchmarking capabilities of \EuroCropsML (Zenodo version 8) by pre-training a state-of-the-art transformer-encoder architecture with sinusoidal positional encoding as introduced by Vasvani, A. et al.\cite{vaswani_vanillatransformer} on both use cases outlined in \cref{sec:benchmkaring_tasks} in the Methods section for a maximum of \num{150} epochs.
The pre-training was terminated if the validation loss did not decrease for more than 15 epochs (early stopping).
Following the pre-training phase, the model's classification head was reset and the model was fine-tuned on the Estonia data in a 1-, 5-, 10-, 20-, 100-, 200-, and 500-shot setting (\cf \cref{sec:benchmkaring_tasks} in the Methods section). 
Furthermore, we propose a baseline comparison in the form of a randomly initialized transformer model, which has not undergone any pre-training before the aforementioned fine-tuning.
The models underwent fine-tuning for a maximum of \num{200} epochs, with validation occurring after each epoch on \num{1000} samples of the validation set.
Again, the training was stopped in case the validation loss did not decrease for more than five epochs (early stopping). 
All experiments were conducted using the Adam optimizer \cite{kingma2017adammethodstochasticoptimization} with a tuned learning rate and a batch size of \num{128}.

\begin{figure}
    \centering
    \input{images/results}
    \caption{Results of the fine-tuning task, indicating the accuracy (\%) obtained on the test set, as also listed in \cref{tab:results}.}
     \label{fig:results}   
\end{figure}

\begin{table}
    \centering
    \begin{tabular}[t]{@{}lrrrrrrr@{}}
        \toprule
        Benchmark task & \multicolumn{7}{l@{}}{Shots $k$} \\
        \cmidrule(l){2-8} 
        & \multicolumn{1}{l}{1} & \multicolumn{1}{l}{5} & \multicolumn{1}{l}{10} & \multicolumn{1}{l}{20} & \multicolumn{1}{l}{100} & \multicolumn{1}{l}{200} & \multicolumn{1}{l@{}}{500}\\
        \cmidrule(r){1-1} \cmidrule(lr){2-2} \cmidrule(lr){3-3} \cmidrule(lr){4-4} \cmidrule(lr){5-5} \cmidrule(lr){6-6} \cmidrule(lr){7-7} \cmidrule(l){8-8}
        No pre-training & \num{0.04} & \num{0.10} & \num{0.28}  & \num{0.41}  & \num{0.45} & \num{0.50}  & \num{0.57}  \\
        Pre-training on LV & \textbf{\num{0.17}}  & \textbf{\num{0.29}}  & \textbf{\num{0.33}} & \textbf{\num{0.44}} & \textbf{\num{0.53}} & \textbf{\num{0.59}} & \textbf{\num{0.66}} \\
        Pre-training on LV+PT & \num{0.06} & \num{0.12} & \num{0.12} & \num{0.18} & \num{0.48} & \num{0.51}  & \num{0.58} \\
        \bottomrule
    \end{tabular}
    \caption{Results of the fine-tuning task, indicating the accuracy (\%) obtained on the test set, as also visualized in \cref{fig:results}.}
    \label{tab:results}
\end{table}

\Cref{fig:results} and \cref{tab:results} show the classification accuracies on the test set.
The model pre-trained on Latvian data demonstrates superior performance across all settings when compared to the two other approaches.
Although the incorporation of data from Portugal generally contributes to the enhancement of the model relative to the baseline, the improvement is not as pronounced as when only pre-training on Latvian data. Furthermore, for $k \in \{10, 20\}$, the randomly initialized model in fact outperforms pre-training with data from Portugal.
This indicates that pre-training on Latvia exploits more characteristics relevant for the subsequent fine-tuning task.
Notably, the analogous climate zones and shared super-classes, as illustrated in \cref{fig:eurocropsmlmap}, render pre-training on Latvia a valuable approach.

\section*{Usage Notes}
\label{sec:usage_notes}
The \EuroCropsML dataset is accompanied by the associated open-source Python package \texttt{\small eurocropsml}. 
It provides convenient access to the data and facilitates its use and integration within the \texttt{\small PyTorch} framework for machine learning.
The package is available on the \emph{Python Package Index (PyPI)} and can be installed via:
\begin{lstlisting}[language=bash]
$ python -Im pip install eurocropsml
\end{lstlisting}
The package is divided into two main sub-packages: The first is concerned with the acquisition of raw data (\cf \cref{sec:level1_data_collection} in the Methods section). 
The second is used to build ready-to-use machine learning datasets from the obtained raw data, \cf \cref{sec:level2_data_processing,sec:benchmkaring_tasks} in the Methods section.
Both sub-packages are accessible through the \emph{command-line interface (CLI)} that allows users to conveniently interact with the package.

For assistance with available commands and other options, we refer to
\begin{lstlisting}[language=bash]
$ eurocropsml-cli --help
\end{lstlisting}
The package can be utilized for the collection and processing of additional data, either on EOLab or any other platform that has Sentinel-2 \texttt{\small .SAFE} files available. Moreover, the CLI provides users with the ability to create their own benchmark tasks, \cf \cref{sec:benchmkaring_tasks} in the Methods section.

The settings depend on several \texttt{\small .yaml}-configuration files located inside the \texttt{\small eurocropsml.configs} module.
When making adjustments to the settings, the user has the option to either create a completely new custom configuration or to override individual parameter values of existing configurations via the CLI.

In the following sections, we will introduce both sub-packages with their most important commands.
A more detailed documentation including the full \emph{application programming interfaces (API)} and CLI commands reference is available at \url{https://eurocropsml.readthedocs.io/en/latest}.

\subsection{Level I: Acquisition pipeline}
\label{sec:level1_usage}
As described in \cref{sec:level1_data_collection} in the Methods section, the acquisition pipeline is used to procure and process Sentinel-2 observations for a given country.
The following command offers further assistance on how to interact with the sub-package:
\begin{lstlisting}[language=bash]
$ eurocropsml-cli acquisition eurocrops --help
\end{lstlisting}
There are pre-defined configuration files available in the \texttt{\small eurocropsml.configs.acquisition.cfg} module for each of the regions of interest (ROI) countries Estonia, Latvia, and Portugal.
The default configuration collects data for Portugal.
To collect data for another country instead, \eg Estonia, the default value can be modified:
\begin{lstlisting}
$ eurocropsml-cli acquisition eurocrops <COMMAND> cfg=estonia
\end{lstlisting}
To process additional countries, it is recommended to create a new configuration file for each country and select it from the command line as shown above.
However, individual changes to an existing configuration, such as collecting data only for the month of January, can also be achieved via the CLI: 
\begin{lstlisting}
$ eurocropsml-cli acquisition eurocrops <COMMAND> cfg.country_config.months="[1,1]" 
\end{lstlisting}

\subsection{Level II: Dataset pipeline}
\label{sec:level2_usage}
The dataset pipeline further processes the raw \EuroCropsML dataset (\cf \cref{sec:level2a}) and provides pre-defined Python classes to ease the use of the dataset (\cf \cref{sec:level2b}).
The whole pipeline is to some extent customizable to the needs of the user. 
The latest version of the dataset can be downloaded from our official Zenodo record \cite{reuss_macdonald_eurocropsml_2024} via:
\begin{lstlisting}[language=bash]
$ eurocropsml-cli datasets eurocrops download
\end{lstlisting}
The following command provides additional support on how to use the dataset sub-package:
\begin{lstlisting}[language=bash]
$ eurocropsml-cli datasets eurocrops --help
\end{lstlisting}

\subsubsection{Level II-A Pre-processing}
\label{sec:level2a}
The settings for the data pre-processing stage are specified in the  \texttt{\small .yaml} configuration file that is located within the \texttt{\small eurocropsml.configs.dataset.preprocess} module.
Individual changes to an existing parameter value can again be made directly through the CLI.
The following command can be used in order to disable the removal of cloudy observations:
\begin{lstlisting}[language=bash]
$ eurocropsml-cli datasets eurocrops <COMMAND> preprocess.filter_clouds=false
\end{lstlisting}
To adjust the lower and upper thresholds $t_1$ and $t_2$ which determine whether an observation is classified as cloudy or non-cloudy, the user can customize the values.
For example, to set the lower bound to \num{0.04} and the upper bound to \num{0.2}:
\begin{lstlisting}[language=bash]
$ eurocropsml-cli datasets eurocrops <COMMAND> preprocess.band4_t1=0.04 preprocess.band4_t2=0.2
\end{lstlisting}
If multiple customizations are required, it is advisable to create a new custom configuration \texttt{\small.yaml} file, for instance \texttt{\small \path{eurocropsml/configs/dataset/preprocess/custom_config.yaml}} and to select it via the command line:
\begin{lstlisting}[language=bash]
$ eurocropsml-cli datasets eurocrops <COMMAND> preprocess=custom_config
\end{lstlisting}

\subsubsection{Level II-B Dataset utilization}
\label{sec:level2b}
The \EuroCropsML dataset allows users to customize options for various crop type classification scenarios, making it suitable for a range of benchmarking applications. Adjustments can be made by creating a custom split configuration within the \texttt{\small eurocropsml.configs.dataset.split} module or by modifying the parameters of existing configurations. 
Detailed split configuration parameters are listed in  \cref{tab:splitconfig}.

In order to customize a transfer learning experiment, there are two class configuration parameters \texttt{\small pretrain\_classes} and \texttt{\small finetune\_classes} as well as two region parameters \texttt{\small pretrain\_regions} and \texttt{\small finetune\_regions}.
They can be specified by providing collections of key-value pairs (dictionaries), respectively. 
The keys can be \texttt{\small region}, \texttt{\small regionclass}, or \texttt{\small class}. Each key refers to a different transfer learning use case, \cf \cref{tab:usecases}. 
The valid values for the class configurations are the HCAT class labels as used by the \EuroCrops reference data, while for the region configurations they are the NUTS region IDs for levels \num{0}--\num{3}.

\begin{table}[t]
  \centering
  \begin{tabular}[t]{@{}p{.20\linewidth}p{.75\linewidth}@{}}
    \toprule
    \textbf{Parameter} & \textbf{Definition} \\
    \midrule
    \texttt{\small base\_name} &  Base name of the split configuration, used when creating and saving the splits.\\
    \texttt{\small data\_dir} &  Folder inside the data directory where pre-processed data is stored.\\
    \texttt{random\_seed} &  Random seed used for generating training-testing-splits and further random numbers.\\
    \texttt{\small num\_samples} &  Number of samples per class used for the fine-tuning subsets. The default will create the shots described in \cref{sec:benchmkaring_tasks} in the Methods section for the training set. 
    It will sample \num{1000} samples for validation and keep all available data from the test set.\\
    \texttt{\small meadow\_class} &  Class that represents the \meadow label. 
    If provided, then this class will be downsampled to the median frequency of all other classes for the pre-training dataset since it represents an imbalanced majority class, as visualized in \cref{fig:numbersamples}.\\
    \texttt{\small pretrain\_classes} & Classes that make up the pre-train dataset.\\
    \texttt{\small finetune\_classes} & Classes that make up the fine-tune dataset.\\
    \texttt{\small pretrain\_regions} & Regions that make up the pre-train dataset.\\
    \texttt{\small finetune\_regions} & Regions that make up the fine-tune dataset.\\
    \bottomrule
  \end{tabular} 
  \caption{Detailed overview and definition of parameters in the split configuration of the \texttt{\small eurocropsml} package.}
    \label{tab:splitconfig}
\end{table}

\begin{table}
  \centering
  \begin{tabular}[t]{@{}p{.15\linewidth}p{.80\linewidth}@{}}
    \toprule
    \textbf{Parameter} & \textbf{Use Case} \\
    \midrule
    \texttt{\small region} &  The data set is divided into pre-training and fine-tuning subsets based on NUTS regions, with three scenarios:
    \textbf{Complete Class Overlap:} All classes between regions overlap, allowing fine-tuning of pre-trained models.
\textbf{Partial Class Overlap:} Some classes overlap between regions, so overlapping classes appear in both pre-training and fine-tuning sets (\cf \cref{sec:benchmkaring_tasks} in the Methods section).
\textbf{No Class Overlap:} No classes overlap between regions, requiring the classifier model to adapt to new classes during fine-tuning.\\
    \texttt{\small regionclass} & The dataset is divided into pre-training and fine-tuning subsets by NUTS regions and subsequently by classes. 
    This setting can be used when classes overlap (partially) between regions, but the pre-train and fine-tune sets should contain different regions and classes.\\
    \texttt{\small class} & The pre-training and fine-tuning subsets are based solely on classes, focusing on knowledge transfer between different sets of classes, not regions. There are two scenarios:
\textbf{Partial Overlap:} Overlapping classes appear in both pre-training and fine-tuning datasets.
\textbf{No Overlap:} The classifier model must adapt to entirely new classes during fine-tuning. \\
    \bottomrule
  \end{tabular} 
    \caption{Detailed description of transfer learning use cases.}
    \label{tab:usecases}
\end{table}

\section*{Code availability}
\label{sec:code_availability}
The code used to generate the dataset is open source and available \url{https://github.com/dida-do/eurocropsml}. It contains pre-defined configuration files which can be used in order to re-process the current dataset.

\section*{Acknowledgements}
The project is funded by the German Federal Ministry for Economics and Climate Action on the basis of a decision by the German Bundestag under funding reference 50EE2007B (J.R. and M.K.), 50EE2007A (J.M., S.B. and L.R.), and 50EE2105 (M.K.). 
S.B. acknowledges support by the Swiss National Science Foundation (SNF) under grant PZ00P2 216019.

\section*{Author contributions statement}
J.R., J.M., S.B., L.R., and M.K. contributed to the study conception and design. 
J.R. performed the collection, analysis, and validation of the data. 
J.R., and J.M. created the accompanied open-source Python package.
J.R., J.M., and S.B. conceived and conducted the experiments and analyzed the results.
All authors contributed to the article, reviewed the manuscript, and approved the submitted version.

\section*{Competing interests}
The authors declare no competing interests.

\end{document}

%% file: images/area_histograms.tex
\begin{tikzpicture}

\definecolor{indianred}{RGB}{196,78,82}
\definecolor{lavender}{RGB}{234,234,242}
\definecolor{mediumseagreen}{RGB}{85,168,104}
\definecolor{peru}{RGB}{221,132,82}
\definecolor{steelblue}{RGB}{76,114,176}

\pgfplotstableread[col sep = comma]{images/area-histograms.csv}\table

\pgfplotsset{
BarPlot/.style={
    axis background/.style={fill=lavender},
    axis line style={white},
    font=\footnotesize,
    ybar,
    bar width=0.25,
    bar shift=-0.125,
    width=0.99\linewidth,
    height=3cm,
    log origin=infty,
    ymax=1000000,
    ytick={1,1000,1000000},
    ymajorgrids,
    ymajorticks=true,
    ytick style={color=white},
    y grid style={white},
    ylabel=\# parcels,
    xtick=data,
    xmin=0.0,
    xmax=13.25,
    minor xtick={0.00,0.25,0.50,0.75,...,13.25},
    xmajorticks=false,
    xtick pos=left,
    xmajorticks=false,
    xtick=false,
    ytick=false,
}
}

\begin{groupplot}[group style={group size=1 by 4, vertical sep=1em}]
\nextgroupplot[
BarPlot,
ymode=log,
log base y=10,
]
\addplot[draw=none,fill=steelblue, opacity=0.75] table [x=kilometerbins, y=LV] {\table};
\node at (axis description cs:0.85,0.8) {\scriptsize \# parcels: \num{431143}};

\nextgroupplot[
BarPlot,
ymode=log,
log base y=10,
]
\addplot[draw=none,fill=peru, opacity=0.75] table [x=kilometerbins, y=EE] {\table};
\node at (axis description cs:0.85,0.8) {\scriptsize \# parcels: \num{175906}};

\nextgroupplot[
BarPlot,
ymode=log,
log base y=10,
]
\addplot[draw=none,fill=mediumseagreen, opacity=0.75] table [x=kilometerbins, y=PT] {\table};
\node at (axis description cs:0.85,0.8) {\scriptsize \# parcels: \num{99634}};
\nextgroupplot[
BarPlot,
ymode=log,
log base y=10,
xtick={0.0,1.0,2.0,...,13.25},
xmajorticks=true,
xlabel=parcel area (\SI{}{\kilo\meter\squared}),
legend style={
    legend columns=4,
   /tikz/every even column/.append style={column sep=1em},
    draw=none,
    fill=none,
    at={(0.5,-0.8)},
    anchor=north,
}
]
\addplot[draw=none,fill=indianred, opacity=0.75, forget plot] table [x=kilometerbins, y=total] {\table};
\node at (axis description cs:0.85,0.8) {\scriptsize \# parcels: \num{706683}};

\addlegendimage{draw=none,fill=steelblue, opacity=0.75}
\addlegendimage{draw=none,fill=peru, opacity=0.75}
\addlegendimage{draw=none,fill=mediumseagreen, opacity=0.75}
\addlegendimage{draw=none,fill=indianred, opacity=0.75}
\addlegendentry{Latvia}
\addlegendentry{Estonia}
\addlegendentry{Portugal}
\addlegendentry{Total}

\end{groupplot}

\node[font=\bfseries, anchor=north] at ($(current bounding box.south)+(0.5,-0.25cm)$) {\textbf{(a)}};

\end{tikzpicture}

%% file: images/class_histograms.tex
\begin{tikzpicture}

\definecolor{indianred}{RGB}{196,78,82}
\definecolor{lavender}{RGB}{234,234,242}
\definecolor{mediumseagreen}{RGB}{85,168,104}
\definecolor{peru}{RGB}{221,132,82}
\definecolor{steelblue}{RGB}{76,114,176}

\pgfplotstableread[col sep = comma]{images/class-histograms.csv}\table

\pgfplotsset{
BarPlot/.style={
    axis background/.style={fill=lavender},
    axis line style={white},
    font=\footnotesize,
    ybar,
    bar width=0.5,
    width=0.99\linewidth,
    height=3cm,
    log origin=infty,
    ymax=1000000,
    ytick={1,1000,1000000},
    ymajorgrids,
    ymajorticks=true,
    ytick style={color=white},
    y grid style={white},
    ylabel=\# parcels,
    xtick=data,
    xmin=-0.5,
    xmax=175.5,
    xticklabel style={text width=1.5cm, rotate=90, align=center},
    xmajorticks=false,
    tick align=outside,
}
}

\begin{groupplot}[group style={group size=1 by 4, vertical sep=1em}]
\nextgroupplot[
BarPlot,
ymode=log,
log base y=10,
]
\addplot[draw=none,fill=steelblue] table [x expr=\coordindex, y=LV] {\table};
\node at (axis description cs:0.85,0.8) {\scriptsize \# crop classes: 103};

\nextgroupplot[
BarPlot,
ymode=log,
log base y=10,
]
\addplot[draw=none,fill=peru] table [x expr=\coordindex, y=EE] {\table};
\node at (axis description cs:0.85,0.8) {\scriptsize \# crop classes: 127};

\nextgroupplot[
BarPlot,
ymode=log,
log base y=10,
]
\addplot[draw=none,fill=mediumseagreen] table [x expr=\coordindex, y=PT] {\table};
\node at (axis description cs:0.85,0.8) {\scriptsize \# crop classes: 79};

\nextgroupplot[
BarPlot,
ymode=log,
log base y=10,
xlabel=crop class,
legend style={
    legend columns=4,
   /tikz/every even column/.append style={column sep=1em},
    draw=none,
    fill=none,
    at={(0.5,-0.75)},
    anchor=north,
}
]
\addplot[draw=none,fill=indianred, forget plot] table [x expr=\coordindex, y=total] {\table};
\node at (axis description cs:0.85,0.8) {\scriptsize \# crop classes: 176};

\addlegendimage{draw=none,fill=steelblue, opacity=0.75}
\addlegendimage{draw=none,fill=peru, opacity=0.75}
\addlegendimage{draw=none,fill=mediumseagreen, opacity=0.75}
\addlegendimage{draw=none,fill=indianred, opacity=0.75}
\addlegendentry{Latvia}
\addlegendentry{Estonia}
\addlegendentry{Portugal}
\addlegendentry{Total}
\end{groupplot}

\node[font=\bfseries, anchor=north] at ($(current bounding box.south)+(0.5,-0.25cm)$) {\textbf{(b)}};

\end{tikzpicture}

%% file: images/zenodo/zenodorawdata.tex
\begin{forest}
  pic dir tree,
  pic root,
  for tree={
    directory,
  },
    [raw\_data
        [geometries
            [Estonia.geojson, file]
            [Latvia.geojson, file]
            [Portugal.geojson, file]
        ]
        [labels 
            [Estonia\_labels.parquet, file]
            [Latvia\_labels.parquet, file]
            [Portugal\_labels.parquet, file]
        ]
        [Estonia.parquet, file
        ]
        [Latvia.parquet, file
        ]
        [Portugal.parquet, file
        ]
    ]
\end{forest}

%% file: images/zenodo/zenodopreprocess.tex
\begin{forest}
  pic dir tree,
  pic root,
  for tree={
    directory,
  },
    [preprocess
        [EE009\_21087530\_3302000000.npz, file]
        [EE004\_20704276\_3301030000.npz, file]
        [EE001\_21796159\_3301090300.npz, file]
        [..., file]
        [LV005\_12799160\_3301010402.npz, file]
        [LV003\_12815133\_3301090303.npz, file]
        [LV008\_12461648\_3302000000.npz, file]
        [..., file]
        [PT16E\_756275\_3301010500.npz, file]
        [PT11E\_1753790\_3303120000.npz, file]
        [PT16D\_306794\_3301990000.npz, file]
        [..., file]
    ]
\end{forest}

%% file: images/zenodo/zenodosplit.tex
\begin{forest}
  pic dir tree,
  pic root,
  for tree={
    directory,
  },
    [splits
        [latvia\_portugal\_vs\_estonia
            [counts
                [finetune
                    [region\_split.csv, file]
                ]
                [pretrain
                    [region\_split.csv, file]
                ]
            ]
            [finetune
                [region\_split\_1.json, file]
                [region\_split\_5.json, file]
                [..., file]
            ]
            [meta
                [region\_split.csv, file]
            ]
            [pretrain
                [region\_split.csv, file]
            ]
        ]
        [latvia\_portugal\_vs\_estonia
            [..., file]
        ]
    ]
\end{forest}

%% file: images/NDVI/NDVI_LV.tex
\begin{tikzpicture}

\definecolor{lavender}{RGB}{234,234,242}
\definecolor{lightblue}{RGB}{100, 181, 205}
\definecolor{steelblue}{RGB}{76,114,176}

\pgfplotstableread[col sep = comma]{images/NDVI/LV_winter_common_soft_wheat.csv}\lvtable
\pgfplotstableread[col sep = comma]{images/NDVI/LV_spring_common_soft_wheat.csv}\lvspringtable

\pgfplotsset{
    LinePlot/.style={
        axis background/.style={fill=lavender},
        axis line style={white},
        font=\footnotesize,
        ymajorgrids,
        ymajorticks=true,
        ymax=1.0,
        ytick={-0.2,0.0,0.2,0.4,0.6,0.8},
        ytick style={color=white},
        y grid style={white},
        xmin=1,
        xmax=366,
        xtick={32,91,152,213,274,335},
        xticklabels={Feb,April,Jun,Aug,Oct,Dec},
        xmajorticks=true,
        xtick pos=left,
        tick align=outside,
        line width=1pt,
    }
}

\begin{groupplot}[group style={group size=1 by 1}]
\nextgroupplot[
LinePlot,
ylabel=NDVI,
legend style={
    legend columns=1,
   /tikz/every even column/.append style={column sep=1em},
    draw=none,
    fill=none,
    at={(axis cs: 360,0.965)},
    anchor=north east,
}
]
\addplot[draw=steelblue,fill=none] table [x=Date, y=winter_common_soft_wheat] {\lvtable};
\addplot[draw=lightblue,fill=none,dashed] table [x=Date, y=spring_common_soft_wheat] {\lvspringtable};;
\addlegendentry{\texttt{\small winter common soft wheat}}
\addlegendentry{\texttt{\small spring common soft wheat}}

\end{groupplot}

\end{tikzpicture}

%% file: images/NDVI/NDVI_EE.tex
\begin{tikzpicture}
\definecolor{lavender}{RGB}{234,234,242}
\definecolor{peru}{RGB}{221,132,82}
\definecolor{lightperu}{RGB}{204, 185, 116}
\pgfplotstableread[col sep = comma]{images/NDVI/EE_winter_common_soft_wheat.csv}\eetable
\pgfplotstableread[col sep = comma]{images/NDVI/EE_spring_common_soft_wheat.csv}\eespringtable
\pgfplotsset{
    LinePlot/.style={
        axis background/.style={fill=lavender},
        axis line style={white},
        font=\footnotesize,
        ymajorgrids,
        ymajorticks=true,
        ymax=1.0,
        ytick={-0.2,0.0,0.2,0.4,0.6,0.8},
        ytick style={color=white},
        y grid style={white},
        xmin=1,
        xmax=366,
        xtick={32,91,152,213,274,335},
        xticklabels={Feb,April,Jun,Aug,Oct,Dec},
        xmajorticks=true,
        xtick pos=left,
        tick align=outside,
        line width=1pt,
    }
}
\begin{groupplot}[group style={group size=1 by 1}]
\nextgroupplot[LinePlot,
legend style={
    legend columns=1,
   /tikz/every even column/.append style={column sep=1em},
    draw=none,
    fill=none,
    at={(axis cs: 360,0.965)},
    anchor=north east,
}
]
\addplot[draw=peru,fill=none] table [x=Date, y=winter_common_soft_wheat] {\eetable};
\addplot[draw=lightperu,fill=none,dashed] table [x=Date, y=spring_common_soft_wheat] {\eespringtable};
\addlegendentry{\texttt{\small winter common soft wheat}}
\addlegendentry{\texttt{\small spring common soft wheat}}
\end{groupplot}
\end{tikzpicture}

%% file: images/NDVI/NDVI_PT.tex
\begin{tikzpicture}

\definecolor{lavender}{RGB}{234,234,242}
\definecolor{mediumseagreen}{RGB}{85,168,104}

\pgfplotstableread[col sep = comma]{images/NDVI/PT_common_soft_wheat.csv}\pttable

\pgfplotsset{
    LinePlot/.style={
        axis background/.style={fill=lavender},
        axis line style={white},
        font=\footnotesize,
        ymajorgrids,
        ymajorticks=true,
        ymax=1.0,
        ytick={-0.2,0.0,0.2,0.4,0.6,0.8},
        ytick style={color=white},
        y grid style={white},
        xmin=1,
        xmax=366,
        xtick={32,91,152,213,274,335},
        xticklabels={Feb,April,Jun,Aug,Oct,Dec},
        xmajorticks=true,
        xtick pos=left,
        tick align=outside,
        line width=1pt,
    }
}

\begin{groupplot}[group style={group size=1 by 1}]
\nextgroupplot[LinePlot,
legend style={
    legend columns=1,
   /tikz/every even column/.append style={column sep=1em},
    draw=none,
    fill=none,
    at={(axis cs: 360,0.965)},
    anchor=north east,
}
]
\addplot[draw=mediumseagreen,fill=none] table [x=Date, y=common_soft_wheat] {\pttable};
\addlegendentry{\texttt{\small common soft wheat}}

\end{groupplot}

\end{tikzpicture}

%% file: images/reflectance/reflectance.tex
\begin{tikzpicture}

\definecolor{lavender}{RGB}{234,234,242}
\definecolor{royalblue}{RGB}{65, 105, 225}
\definecolor{darkorange}{RGB}{255, 140, 0}
\definecolor{green}{RGB}{0, 128, 0}
\definecolor{red}{RGB}{255, 0, 0}
\definecolor{purple}{RGB}{128, 0, 128 }
\definecolor{brown}{RGB}{150, 75, 0} 
\definecolor{violet}{RGB}{238, 130, 238}
\definecolor{gray}{RGB}{128, 128, 128} 
\definecolor{yellowgreen}{RGB}{154, 205, 50} 
\definecolor{cyan}{RGB}{0, 255, 255}
\definecolor{olivedrab}{RGB}{107, 142, 35}
\definecolor{deeppink}{RGB}{255, 20, 147}
\definecolor{teal}{RGB}{0, 128, 128}

 \definecolor{B01}{rgb}{0.29411764705882354,0,0.5098039215686274}
 \definecolor{B02}{rgb}{0.12156862745098039, 0.4666666666666667, 0.7058823529411765}
 \definecolor{B03}{rgb}{0.17254901960784313, 0.6274509803921569, 0.17254901960784313}
 \definecolor{B04}{rgb}{0.8392156862745098, 0.15294117647058825, 0.1568627450980392}
 \definecolor{B05}{rgb}{0.9186274509803922, 0.9068627450980393, 0.9264705882352941}
 \definecolor{B06}{rgb}{0.9055901576316802, 0.8817377931564783, 0.9369319492502883}
 \definecolor{B07}{rgb}{0.8310342176086121, 0.7243521722414455, 0.854317570165321}
 \definecolor{B08}{rgb}{0.7892502883506344, 0.5782237600922723, 0.7793310265282584}
 \definecolor{B8A}{rgb}{0.8750019223375625, 0.3923875432525951, 0.6878585159554017}
 \definecolor{B09}{rgb}{0.9039600153787005, 0.1590157631680123, 0.5371780084582852}
 \definecolor{B10}{rgb}{0.8028604382929643, 0.0689273356401384, 0.33550173010380624}
 \definecolor{B11}{rgb}{0.590803537101115, 0.0, 0.2588696655132641}
 \definecolor{B12}{rgb}{0.403921568627451, 0.0, 0.12156862745098039}

\pgfplotstableread[col sep = comma]{images/reflectance/EE_winter_common_soft_wheat_reflectance.csv}\cloudstable
\pgfplotstableread[col sep = comma]{images/reflectance/EE_winter_common_soft_wheat_reflectance_no_clouds.csv}\nocloudstable

\pgfplotsset{
    LinePlot/.style={        
        axis background/.style={fill=lavender},
        axis line style={white},
        font=\footnotesize,
        width=.55\linewidth,
        height=6cm,
        ytick={0.0,0.25,0.5,0.75,1.0},
        ymajorticks=true,
        ymajorgrids,
        ymax=1.25,
        ymin=0,
        ytick style={color=white},
        y grid style={white},
        xmin=1,
        xmax=366,
        xtick={32,91,152,213,274,335},
        xticklabels={Feb,April,Jun,Aug,Oct,Dec},
        xmajorticks=true,
        xtick pos=left,
        tick align=outside,
        line width=.75pt,
        mark=none,
    }
}

\begin{groupplot}[group style={group size=2 by 1, horizontal sep=1em}]
\nextgroupplot[
    LinePlot,
    ylabel=reflectance,
]
\addplot+[mark=none, color=B01, solid, opacity=0.75, forget plot] table[x={Date}, y={0}] {\cloudstable};
\addplot+[mark=none, color=B05, solid, opacity=0.75, forget plot] table[x={Date}, y={4}] {\cloudstable};
\addplot+[mark=none, color=B06, solid, opacity=0.75, forget plot] table[x={Date}, y={5}] {\cloudstable};
\addplot+[mark=none, color=B07, solid, opacity=0.75, forget plot] table[x={Date}, y={6}] {\cloudstable};
\addplot+[mark=none, color=B08, solid, opacity=0.75, forget plot] table[x={Date}, y={7}] {\cloudstable};
\addplot+[mark=none, color=B8A, solid, opacity=0.75, forget plot] table[x={Date}, y={8}] {\cloudstable};
\addplot+[mark=none, color=B09, solid, opacity=0.75, forget plot] table[x={Date}, y={9}] {\cloudstable};
\addplot+[mark=none, color=B10, solid, opacity=0.75, forget plot] table[x={Date}, y={10}] {\cloudstable};
\addplot+[mark=none, color=B11, solid, opacity=0.75, forget plot] table[x={Date}, y={11}] {\cloudstable};
\addplot+[mark=none, color=B12, solid, opacity=0.75, forget plot] table[x={Date}, y={12}] {\cloudstable};
\addplot+[mark=none, color=B04, solid, opacity=0.75, forget plot] table[x={Date}, y={3}] {\cloudstable};
\addplot+[mark=none, color=B03, solid, opacity=0.75, forget plot] table[x={Date}, y={2}] {\cloudstable};
\addplot+[mark=none, color=B02, solid, opacity=0.75, forget plot] table[x={Date}, y={1}] {\cloudstable};
\node[draw, rectangle, thick, rounded corners, anchor=east, opacity=0.8] at (axis description cs:0.99,0.9) {before cloud removal};

\nextgroupplot[
LinePlot,
yticklabel=\empty,
legend style={
    legend columns=7,
   /tikz/every even column/.append style={column sep=1em},
    draw=none,
    fill=none,
    at={(0,-0.15)},
    anchor=north,
}]
\addplot+[mark=none, color=B01, solid, opacity=0.75, forget plot] table[x={Date}, y={0}] {\nocloudstable};
\addplot+[mark=none, color=B05, solid, opacity=0.75, forget plot] table[x={Date}, y={4}] {\nocloudstable};
\addplot+[mark=none, color=B06, solid, opacity=0.75, forget plot] table[x={Date}, y={5}] {\nocloudstable};
\addplot+[mark=none, color=B07, solid, opacity=0.75, forget plot] table[x={Date}, y={6}] {\nocloudstable};
\addplot+[mark=none, color=B08, solid, opacity=0.75, forget plot] table[x={Date}, y={7}] {\nocloudstable};
\addplot+[mark=none, color=B8A, solid, opacity=0.75, forget plot] table[x={Date}, y={8}] {\nocloudstable};
\addplot+[mark=none, color=B09, solid, opacity=0.75, forget plot] table[x={Date}, y={9}] {\nocloudstable};
\addplot+[mark=none, color=B10, solid, opacity=0.75, forget plot] table[x={Date}, y={10}] {\nocloudstable};
\addplot+[mark=none, color=B11, solid, opacity=0.75, forget plot] table[x={Date}, y={11}] {\nocloudstable};
\addplot+[mark=none, color=B12, solid, opacity=0.75, forget plot] table[x={Date}, y={12}] {\nocloudstable};
\addplot+[mark=none, color=B04, solid, opacity=0.75, forget plot] table[x={Date}, y={3}] {\nocloudstable};
\addplot+[mark=none, color=B03, solid, opacity=0.75, forget plot] table[x={Date}, y={2}] {\nocloudstable};
\addplot+[mark=none, color=B02, solid, opacity=0.75, forget plot] table[x={Date}, y={1}] {\nocloudstable};
\node[draw, rectangle, thick, rounded corners, anchor=east, opacity=0.8] at (axis description cs:0.99,0.9) {after cloud removal};

\addlegendimage{draw=B01,fill=B01}
\addlegendimage{draw=B02,fill=B02}
\addlegendimage{draw=B03,fill=B03}
\addlegendimage{draw=B04,fill=B04}
\addlegendimage{draw=B05,fill=B05}
\addlegendimage{draw=B06,fill=B06}
\addlegendimage{draw=B07,fill=B07}
\addlegendimage{draw=B08,fill=B08}
\addlegendimage{draw=B8A,fill=B8A}
\addlegendimage{draw=B09,fill=B09}
\addlegendimage{draw=B10,fill=B10}
\addlegendimage{draw=B11,fill=B11}
\addlegendimage{draw=B12,fill=B12}
\addlegendentry{$\rho_{B01}$}
\addlegendentry{$\rho_{B02}$}
\addlegendentry{$\rho_{B03}$}
\addlegendentry{$\rho_{B04}$}
\addlegendentry{$\rho_{B05}$}
\addlegendentry{$\rho_{B06}$}
\addlegendentry{$\rho_{B07}$}
\addlegendentry{$\rho_{B08}$}
\addlegendentry{$\rho_{B8A}$}
\addlegendentry{$\rho_{B09}$}
\addlegendentry{$\rho_{B10}$}
\addlegendentry{$\rho_{B11}$}
\addlegendentry{$\rho_{B12}$}

\end{groupplot}

\end{tikzpicture}

%% file: images/results.tex
\begin{tikzpicture}

\definecolor{indianred}{RGB}{196,78,82}
\definecolor{lavender}{RGB}{234,234,242}
\definecolor{mediumseagreen}{RGB}{85,168,104}
\definecolor{peru}{RGB}{221,132,82}
\definecolor{steelblue}{RGB}{76,114,176}

\pgfplotsset{
LinePlot/.style={
    axis background/.style={fill=lavender},
    axis line style={white},
    font=\footnotesize,
    width=0.75\linewidth,
    height=4cm,
    ymin=0,
    ymax=0.75,
    ytick={0.0,0.25,0.5,0.75},
    yticklabels={0.00,0.25,0.50,0.75},
    ylabel style={font=\sffamily\normalsize},
    yticklabel style={font=\sffamily\normalsize}, 
    ymajorgrids,
    ymajorticks=true,
    ytick style={color=white},
    y grid style={white},
    ylabel=accuracy (\%),
    xlabel=shots k,
    xtick={1,5,10,20,100,200,500},
    xticklabels={1,5,10,20,100,200,500},
    xlabel style={font=\sffamily\normalsize},
    xticklabel style={font=\sffamily\normalsize}, 
    xmin=1,
    xmax=500,
    xmajorticks=true,
    tick align=outside,
    xtick pos=left,
}
}

\begin{axis}[
LinePlot,
xmode=log,
log base x=10,
legend style={
    legend columns=1,
    legend cell align=left,
   /tikz/every even column/.append style={column sep=1em},
    draw=none,
    fill=none,
    at={(1.05,0.5)},
    anchor=west,
    font=\sffamily\normalsize,
}
]
\addplot[mark=*,peru,line width=1.5] plot coordinates {
    (1,0.04) (5,0.10) (10,0.28) (20,0.41) (100,0.45) (200,0.50) (500,0.57)
};
\addplot[mark=diamond*,steelblue,line width=1.5] plot coordinates {
    (1,0.17) (5,0.29) (10,0.33) (20,0.44) (100,0.53) (200,0.59) (500,0.66)
};
\addplot[mark=square*,mediumseagreen,line width=1.5] plot coordinates {
    (1,0.06) (5,0.12) (10,0.12) (20,0.18) (100,0.48) (200,0.51) (500,0.58)
};

\addlegendentry{No pre-training}
\addlegendentry{Pre-training on LV}
\addlegendentry{Pre-training on LV+PT}

\end{axis}

\end{tikzpicture}

%% file: main.bbl
\begin{thebibliography}{99}
    \bibitem{land-use} Ritchie, H. \& Roser, M. Land Use. \textit{Our World in Data} \url{https://ourworldindata.org/land-use} (2024).
    \bibitem{turkoglu_cropmapping} Turkoglu, M. O. et al. Crop mapping from image time series: Deep learning with multi-scale label hierarchies. Remote. Sens. Environ. \textbf{264}, 112603 \url{https://doi.org/10.1016/j.rse.2021.112603} (2021).
    \bibitem{DENETHOR} Kondmann, L. et al. DENETHOR: The DynamicEarthNET dataset for harmonized, inter-operable, analysis-ready, daily crop monitoring from space. \textit{Proc. Neural Inf. Process. Syst. Track on Datasets and Benchmarks} \url{https://datasets-benchmarks-proceedings.neurips.cc/paper_files/paper/2021/file/5b8add2a5d98b1a652ea7fd72d942dac-Paper-round2.pdf} (2021).
    \bibitem{russwurm_breizhcrops} Rußwurm, M., Pelletier, C., Zollner, M., Lefèvre, S. \& K\"orner, M. BreizhCrops: A Time Series Dataset for Crop Type Mapping. \textit{ISPRS – Int. Arch. Photogramm. Remote Sens. Spatial Inf. Sci.} \textbf{XLIII-B2-2020}, 1545–1551 \url{https://isprs-archives.copernicus.org/articles/XLIII-B2-2020/1545/2020/} (2020).
    \bibitem{Garnot2021PanopticSO} Garnot, V. S. F. \& Landrieu, L. Panoptic segmentation of satellite image time series with convolutional temporal attention networks. \textit{Proc. IEEE/CVF Int. Conf. Comput. Vis.} 4852–4861 \url{https://api.semanticscholar.org/CorpusID:236034332} (2021).
    \bibitem{Dimitris_2021_sen4agrinet} Sykas, D., Papoutsis, I. \& Zografakis, D. Sen4AgriNet: A harmonized multi-country, multi-temporal benchmark dataset for agricultural earth observation machine learning applications. \textit{Proc. IEEE Int. Geosci. Remote Sens. Symp. 5830–5833} \url{https://doi.org/10.1109/IGARSS47720.2021.9553603} (2021).
    \bibitem{tseng_cropharvest_2021} Tseng, G., Zvonkov, I., Nakalembe, C. L. \& Kerner, H. CropHarvest: A global dataset for crop-type classification. \textit{Proc. Neural Inf. Process. Syst. Track on Datasets and Benchmarks}. Preprint at \url{https://openreview.net/forum?id=JtjzUXPEaCu} (2021).
    \bibitem{LIU2025109364} Liu, S., Wang, L., Zhang, J., and Ding, S. Opposite effect on soil organic carbon between grain and non-grain crops: Evidence from Main Grain Land, China. \textit{Agriculture, Ecosystems \& Environment
Volume 379, 28 February 2025, 109364}.
    \url{https://doi.org/10.1016/j.agee.2024.109364} (2025).
    \bibitem{LI2021106055} Li, Y. and Yang, J. Meta-learning baselines and database for few-shot classification in agriculture. \textit{Computers and Electronics in Agriculture
Volume 182, March 2021, 106055}.
    \url{https://doi.org/10.1016/j.compag.2021.106055} (2021).
    \bibitem{Schmitt_2023} Schmitt, M. et al. There are no data like more data: Datasets for deep learning in earth observation. \textit{IEEE Geosci. Remote. Sens. Mag.} \textbf{11}, 63–97 \url{https://doi.org/10.1109/MGRS.2023.3293459} (2023).
    \bibitem{schneider_eurocrops_zenodo} Schneider, M., Chan, A. \& K\"orner, M. EuroCrops. \textit{Zenodo} \url{https://doi.org/10.5281/zenodo.8229128} (2023).
    \bibitem{schneider_eurocrops_2023} Schneider, M., Schelte, T., Schmitz, F. \& K\"orner, M. EuroCrops: The largest harmonized open crop dataset across the European Union. \textit{Sci. Data} \textbf{10}, 612 \url{https://doi.org/10.1038/s41597-023-02517-0} (2023).
    \bibitem{schneider_eurocrops_2021} Schneider, M., Broszeit, A. \& K\"orner, M. EuroCrops: A pan-european dataset for time series crop type classification. \textit{Proc. Conf. Big Data Space (BiDS)} From Insights to Foresight \url{https://doi.org/10.2760/125905281} (2021).
    \bibitem{eurocrops_github} Schneider, Maja and Chan, Ayshah and K\"orner, Marco. EuroCrops: The official repository for the EuroCrops dataset. \textit{GitHub} \url{https://github.com/maja601/EuroCrops/blob/5c33a1133865ded9ec1a1438fc7708e84a895db5/hcat_core/HCAT3.csv} (2023).
    \bibitem{ESA_Sentinel2_ATBD} Louis, J., Devignot, O. \& Pessiot, L. Sentinel-2 MSI Level-2A algorithms, products, and performance: Algorithm theoretical basis document. S2-PDGS-MPC-ATBD-L2A. Issue 2.10. \textit{European Space Agency} \url{https://step.esa.int/thirdparties/sen2cor/2.10.0/docs/S2-PDGS-MPC-L2A-ATBD-V2.10.0.pdf} (2022).
    \bibitem{reuss_macdonald_eurocropsml_2024} Reuss, J. \& Macdonald, J. EuroCropsML. \textit{Zenodo} \url{https://doi.org/10.5281/zenodo.12168505} (2024).
    \bibitem{vaswani_vanillatransformer} Vaswani, A. et al. Attention is All you Need. \textit{Proc. Adv. Neural Inf. Process. Syst.} \textbf{30} \url{https://proceedings.neurips.cc/paper_files/paper/2017/file/3f5ee243547dee91fbd053c1c4a845aa-Paper.pdf} (2017).
    \bibitem{kingma2017adammethodstochasticoptimization} Kingma, D. P. \& Ba, J. Adam: A method for stochastic optimization. Preprint at \url{https://arxiv.org/abs/1412.6980} (2017).
\end{thebibliography}
